# Optimizing Social Media Annotation of HPV Vaccine Skepticism and Misinformation Using Large Language Models: An Experimental Evaluation of In-Context Learning and Fine-Tuning Stance Detection Across Multiple Models

Luhang Sun[1], Varsha Pendyala[2], Yun-Shiuan Chuang[3,4], Shanglin Yang[3], Jonathan Feldman[5], Andrew Zhao[6], Munmun De Choudhury[5], Sijia Yang[1], and Dhavan Shah[1]

[1] School of Journalism and Mass Communication, University of Wisconsin–Madison

[2] Department of Electrical & Computer Engineering, University of Wisconsin–Madison

[3] Department of Computer Science, University of Wisconsin–Madison

[4] Department of Psychology, University of Wisconsin–Madison

[5] School of Interactive Computing, Georgia Institute of Technology

[6] Institute of People and Technology, Georgia Institute of Technology

# Optimizing Social Media Annotation of HPV Vaccine Skepticism and Misinformation Using Large Language Models: An Experimental Evaluation of In-Context Learning and Fine-Tuning Stance Detection Across Multiple Models

**Abstract**

This paper leverages large-language models (LLMs) to experimentally determine optimal strategies for scaling up social media content annotation for stance detection on HPV vaccine-related tweets. We examine both conventional fine-tuning and emergent in-context learning methods, systematically varying strategies of prompt engineering across widely used LLMs and their variants (e.g., GPT4, Mistral, and Llama3, etc.). Specifically, we varied prompt template design, shot sampling methods, and shot quantity to detect stance on HPV vaccination. Our findings reveal that 1) in general, in-context learning outperforms fine-tuning in stance detection for HPV vaccine social media content; 2) increasing shot quantity does not necessarily enhance performance across models; and 3) different LLMs and their variants present differing sensitivity to in-context learning conditions. We uncovered that the optimal in-context learning configuration for stance detection on HPV vaccine tweets involves six stratified shots paired with detailed contextual prompts. This study highlights the potential and provides an applicable approach for applying LLMs to research on social media stance and skepticism detection.

*Word count: 7,437 excluding references and appendices*

## 1. Introduction

Computational social science research often involves applying computer-assisted techniques to process large-scale data resources to understand human behaviors on a societal scale (Lazer et al., 2009, 2020). Researchers have increasingly recommended integrating AI-assisted techniques and LLMs into the workflow of computational social science research, which can reduce significant costs in terms of time, human labor, financial resources, and technical expertise (Törnberg, 2024; Ziems et al., 2024). Indeed, the rise of LLMs heralds a "paradigm shift" in computational social science, especially for applications to repetitive, expertise-laden, and time-consuming tasks, such as latent feature annotation (also referred to as "labeling" or "coding"), which are essential for training supervised machine learning models. These tools have the potential to revolutionize the study of communication, digital conversation, and media technologies, especially in research domains such as misinformation detection and message framing analysis, which require high levels of accuracy and interpretive skill within specific contexts to identify linguistic patterns and generate reliable categorizations. However, given the nascency of the AI turn in computational social science research and the variety of options to implement LLMs in a research project, evidence is needed to identify the optimal strategy for AI deployment. We tackle this critical gap by systematically documenting the performance of LLMs while varying several key dimensions for implementation in the context of large-scale stance detection for social media posts discussing HPV vaccination.

Specifically, we compared two major approaches to potentially enhance the performance of general-purpose LLMs for specific social science applications including stance detection for social media data—fine-tuning versus in-context learning. Within in-context learning, we further systematically varied the following dimensions: prompt design, number of shots and selection strategies, and variants of LLM models. Our study evaluates the performance of different in-context learning design dimensions and fine-tuning approaches across the LLMs, discusses the implications of specific practices, and suggests avenues for integrating LLMs in computational social science, political communication, and health communication research. Key practices we explore include prompt message design, example shot sampling methods, shot selection rationale, and fine-tuning strategies. Our goal is to provide practical insights and performance evidence that can serve as references for future social science research, potentially accelerating research centering on communication, especially consequential digital conversations taking place over online platforms.

The context of the study is human papillomavirus (HPV) vaccine discussions on Twitter, selected due to the prevalence of vaccine hesitancy and misinformation on social media (Massey et al., 2020), particularly surrounding vaccines with relatively shorter histories, such as the HPV vaccine, which continues to face low uptake (Vraga et al., 2023), as well as the politicization of HPV vaccine policy support (Saulsberry et al., 2019). Accurate stance classification in social media data is particularly crucial in this context, as it enables researchers and stakeholders to identify and understand the linguistic nuances and rhetorical strategies of anti-vaccine sentiments shared over media technologies. Moreover, this research focuses on stance classification of low-credibility content, expanding beyond the narrower notion of explicit misinformation to include vaccine skepticism. A recent large-scale study has demonstrated that social media posts questioning the efficacy of COVID-19 vaccines (i.e., vaccine-skeptical content) far outweighed labeled misinformation in driving vaccine hesitancy among the US public (Allen et al., 2024). Assessing, at scale, the wider range of content questioning HPV vaccine efficacy would allow more targeted health communication interventions and campaigns in our polarized and politicized information environment. Recognizing the challenges of prior supervised machine learning annotation tasks, we aim to demonstrate the efficacy of LLMs and their variants in detecting HPV vaccine stance using both in-context learning experimental design and fine-tuned models.

Our results indicate that GPT-4 Turbo outperforms other frontier large language models in overall performance metrics employing the in-context learning approach. However, increasing the number of shots does not necessarily enhance performance, particularly for GPT-4 Turbo (the top available model at the time of this analysis), where the decline may be attributed to *cognitive overload* in in-context learning. Generally, our findings suggest that in-context learning configurations outperform their fine-tuning counterparts, with models and their variants presenting different levels of sensitivity to the manipulated in-context learning conditions in the experiment. The optimal approach of in-context learning for stance classification of HPV vaccine tweets involves using six stratified shot examples and detailed contextual prompts to guide GPT-4 Turbo to perform effectively as an expert content analyst. These findings highlight both the potential and challenges of integrating LLMs into computational social science research in the domain of health and politics, emphasizing the importance of fine-tuning in-context learning conditions, model-specific adaptations, and human-in-the-loop interventions. We expand the emerging literature (Demszky et al., 2023) on how LLMs could be harnessed effectively to complement human skills in a cooperative fashion within the domain of computational social science.

## 2. Background and literature review

### *2.1. Vaccine hesitancy, distrust, and misinformation in social media discourse*

Vaccine hesitancy has been a global challenge and threat to public health (Bussink-Voorend et al., 2022), particularly for the vaccines with relatively shorter histories, such as the low-uptake HPV vaccination (Vraga et al., 2023). Researchers in public health and health communication have endeavored to study the determinants and interventions of the HPV vaccine and other vaccine hesitancy, as well as the linguistic features of social media discussions around vaccine hesitancy, misinformation, and conspiracy theories (Chen et al., 2020; Ortiz et al., 2019; Puri et al., 2020). Social media has been viewed as a crucial channel for the dissemination of (mis)information and skepticism regarding HPV and other vaccines, as it has become a primary source of health-related information for many members of the public, which may potentially affect their perceptions and intentions regarding vaccination (Dunn et al., 2017; Nan & Madden, 2012; Vraga et al., 2023).

Various empirical studies show that negative attitudes toward vaccination (e.g., concerns about vaccine effectiveness and safety), alongside political ideologies and conspiracy theories, are prevalent around discourses of vaccine hesitancy and misinformation on social media (Dhaliwal & Mannion, 2020; Di Domenico et al., 2022; Massey et al., 2020). Conversely, research finds that greater perceived certainty about the scientific evidence for HPV vaccine is tied to more support for HPV vaccine policies (Saulsberry et al., 2019). Messages expressing vaccine hesitancy and misinformation are often conveyed through personal narratives and anecdotes by individual users on social media (Massey et al., 2020). To understand how individual online users make sense of their beliefs and intentions regarding vaccination, particularly vaccine hesitancy, it is important to examine the actual content people encounter or express on social media. Accordingly, recent studies have examined social media content at scale by integrating various computer-assisted techniques across different vaccination contexts. For instance, Jiang and colleagues (2021) employed both supervised and unsupervised machine learning annotation techniques to understand COVID-19 vaccine skepticism and its ideological differences. They found that online users with conservative leanings were more likely to post messages expressing distrust or conspiracy theories about COVID-19 vaccines. In their study, Jiang and colleagues operationalized the vaccine stance by training human coders to annotate multiple latent message features, such as vaccine favorability, side effects, and distrust. These annotated data were then used to train and fine-tune Bidirectional Encoder Representations from Transformers (BERT) models to classify messages based on these latent features, including "vaccine favorability," a stance detection metric relevant to understanding public sentiment and attitudes toward vaccination. This approach of combining human annotation with machine learning for stance detection is similar to the methods applied in the present study on HPV vaccination discourse, where stance detection serves as a critical tool for identifying in-favor, neutral, or opposing views on vaccination.

### *2.2. Integrating LLMs into text classification*

To better understand the vast and diverse social media landscape surrounding vaccine attitudes and discourses, researchers have turned to computational methods, including dictionary-based approaches (e.g., Himelboim et al., 2019; King et al., 2023; Wang et al., 2019), unsupervised machine learning clustering approaches (e.g., Hwang et al., 2022; Jiang et al., 2021), and supervised machine learning approaches (e.g., Chuang et al., 2023; Piedrahita-Valdés et al., 2021; Sun et al., 2023). These approaches enable researchers to classify textual content for stance detection, sentiment analysis, and topic features, which serve as foundational elements for further analysis and interpretation. Admittedly, supervised machine learning models offer the advantage of being tailored to specific data characteristics and message features of interest, by training classifiers on their own for specific contexts and purposes. However, this process may require considerable time, human labor, financial resources, and technical expertise, as a conventional supervised machine learning approach in computational social science often involves many stages, from data collection, data pre-processing, high-volume human annotations, model training and evaluation.

The recent rise of LLMs in computational social science has the potential to address some of these challenges by enabling more efficient, accurate, and cost-benefit data analysis. Researchers have begun integrating LLMs to streamline feature construction and classification within large-scale datasets across various research contexts (e.g., Heseltine & Clemm von Hohenberg, 2024; Tan et al., 2024). For instance, Ziems et al. (2024) evaluate multiple LLMs for social science applications, recommending the integration of LLMs in annotations and generation tasks due to their high efficiency and cost-effectiveness. Likewise, Törnberg (2024) argues that LLMs outperform supervised classifiers and even expert coders in the context of political social media messages.

Indeed, while the incorporation of LLMs into computational social science research has proven beneficial, due to the low cost, high efficiency, and performance accuracy, it also raises new questions about optimizing these tools for varied research needs. As the number of LLMs and their customization options increase, it is time to scrutinize and refine prompt engineering (i.e., designing prompt content for task guidance), in-context learning (i.e., embedding examples within prompts to provide context for the task), and fine-tuning techniques (i.e., adjusting model parameters for enhanced performance) to improve performance across different tasks and LLMs (Yao et al., 2023). Specifically, prompt quality is likely to impact model performance, particularly in tasks like classifying social media content that may require nuanced interpretive skills. In addition to prompt template *per se*, in-context learning—a component of prompt engineering that embeds relevant examples within prompts—also plays a crucial role in enhancing model performance by providing LLMs with contextual information. Moreover, fine-tuning methods, a conventional yet complementary approach to prompt engineering, offer distinct advantages for integrating LLMs into text classification. Unlike prompt engineering, fine-tuning does not require additional efforts to design prompt templates or provide in-context examples, as it customizes the model's parameters directly to the task. We believe this systematical scrutiny will enable social scientists to optimize the efficiency of LLMs as computational tools in their research.

### ***2.3. Prompt engineering and in-context learning using LLMs***

In-context learning—the ability of LLMs to generalize from a few examples provided within the prompt—has proven effective across a wide variety of language understanding tasks (Dong et al., 2022; Liu et al., 2021). Unlike conventional supervised machine learning, where models often require large amounts of annotated data for training, in-context learning allows models to adapt on the fly to new tasks by leveraging contextual information. This adaptability makes in-context learning particularly valuable for content annotation tasks, where efficient, low-cost solutions are often essential. However, the performance of prompt engineering and in-context learning depends on a combination of factors, such as the prompting template, the selection and number of in-context examples, and the order in which these examples are presented (Zhao et al., 2021).

In the present study, we evaluate the impact of different prompt templates (by varying the depth and detail of the instructions) and different shot selection strategies (by varying the number and sampling method of in-context examples) on model performance. In prior studies, researchers have emphasized the importance of prompt template design, focusing on various features such as structure, specificity, clarity, and language framing within the instructions. Prompt templates may vary across dimensions like task framing, contextualization, and response style, each of which can affect the model's ability to capture task-specific language cues (Ma et al., 2021; Schick & Schütze, 2020a, 2020b). Building on these insights, we investigate whether using detailed prompt templates with high specificity and clear definitions for each level of stance detection will improve model performance across different LLMs. Specifically, the considerations of our prompt design vary in four sub-dimensions, including (a) the role of the LLM, (b) category definitions and details, (c) a wider range of language markers and forms, and (d) the importance of accurate classification.

Shot selection strategies also play a critical role in in-context learning performance: selecting optimal examples to include in the prompt, referred to as "shots," affects the model's understanding of the task. Zero-shot and few-shot learning has shown promise in content annotation tasks, with researchers seeking to optimize the performance of these models through specific efforts at prompt engineering, often focusing on the prompt itself and the selection of shots used for in-context learning (Song et al., 2023). Others have focused on few-shot selection—optimizing strategies for appropriate sample selection—as another important element for in-context learning (An et al., 2023). Optimizing the selection of shots, both the number needed to provide sufficient context to the LLMs and the sampling strategy used to select these shots has drawn research attention. While advanced machine learning methods for selecting in-context examples exist, such as k-NN-based unsupervised retrieval (Liu et al., 2021), they can be computationally demanding and costly to implement. To address this, our study utilizes two straightforward shot selection methods: random sampling and stratified sampling from the annotated training data. Random sampling provides a baseline approach by selecting examples directly from the social media dataset without any systematic/stratified strategy; while stratified sampling ensures that selected examples are stratified based on the three stance levels within the dataset. By comparing these two approaches, we aim to understand the extent to which shot selection impacts the model's accuracy in detecting stances regarding HPV vaccination on social media.

The number of shots, or in-context examples, is another consideration in our study. Agarwal and colleagues (2024) found that increasing the number of shots (i.e., from few-shot to many-shot) yielded significant improvements in model performance across a wide variety of tasks, with more examples allowing the model to override pretraining biases and improve task specificity, asserting that "unlike few-shot learning, many-shot learning is effective at overriding pretraining biases" (p. 1). However, we believe that balancing the benefits of additional examples with computational efficiency remains a critical consideration. Our study explores both few-shot and many-shot, as well as zero-shot as baseline configurations to understand how shot quantity influences the model performance of the stance detection task. In addition, we account for model sensitivity by testing several widely used state-of-the-art LLMs, both proprietary and open-source. Previous studies indicate that different LLMs exhibit varying sensitivity to in-context learning across tasks, such as sentiment analysis (Zhang et al., 2023). This variation can be attributed to differences in model architecture, pretraining data, and parameter count, which affect how models interpret and respond to prompt structures. Therefore, by evaluating multiple

LLMs, our study identifies best practices in prompt engineering that may be generalized across models or tailored to specific model architectures.

### *2.4. Fine-tuning approaches in LLMs*

Recognizing that in-context learning is not the only paradigm to enhance the performance of LLMs for natural language processing and stance classification, our study also takes fine-tuning approaches into consideration. Indeed, as a relatively new strategy, in-context learning was first used to describe the emergent behavior of LLMs only several years ago, following the rapid dilation of LLM data and model size (Brown et al., 2020; Dong et al., 2022). The more conventional and established strategy of adapting LLMs to correspond to a specific downstream task is fine-tuning. Fine-tuning is a process through which changes are introduced to some of the parameters of an LLM while holding the rest constant. By changing only a small fraction of the total model weights, an LLM receives new, and often more specialized, knowledge while maintaining the broad understanding of the model developed during pre-training, which remains accessible to the model through the unaltered weights. Fine-tuning allows a generalist LLM to hone its capabilities to complete a specified task, such as sentiment analysis or misinformation detection.  (Dong et al., 2022; Mosbach et al., 2023).

It is still heavily debated whether fine-tuning modern LLMs is less beneficial than employing in-context learning, especially since in-context learning is often more computationally burdensome during inference, and it is unclear whether one method is wholly better than the other for all tasks. Previous studies have shown that fine-tuned models outperform LLMs employing in-context learning on domain-specific tasks, while other studies have contradicted those very findings (e.g., Liu et al., 2024; Mosbach et al., 2023; Yin et al., 2024). Furthermore, the distinction between the two methods is heavily dependent on various factors, including the architecture of the model, the quality of the dataset, and the specific task assigned to the model (Mosbach et al., 2023; Tekumalla & Banda, 2023). However, few previous studies specifically examined the relative efficacies of fine-tuning versus in-context learning for stance classification within health misinformation contexts, such as HPV vaccine discourse. Thus, our study aims to systematically examine the relative performance of fine-tuned and in-context learning-based models.

Furthermore, there may be other benefits to fine-tuning: given that a fine-tuned model's weights are altered for a specific task, contextualization is not needed during inference, which may lead to increased scalability and computational efficiency (Mosbach et al., 2023; Xia et al., 2024). The efficiency of fine-tuning is heightened if the fine-tuning technique used to modify the pre-trained LLM falls within the family of Parameter-Efficient Fine-Tuning (PEFT) methodologies, which significantly reduce computing time during the fine-tuning process as compared to classical fine-tuning techniques (Xu et al., 2023). By incorporating and comparing in-context learning and fine-tuning, our study assesses the influence of prompt design, shot selection strategies, and fine-tuning methods on the performance of LLMs for stance detection in the context of HPV vaccine social media discourse. Our findings aim to provide practical guidance on optimizing LLM-based text annotation for computational social science research, demonstrating the potential for achieving high accuracy and efficiency in classifying complex and even controversial social media data.

## 3. Methods

### *3.1. Experimental design for prompt engineering and in-context learning*

Our experimental design primarily tests three key dimensions of prompt design and in-context learning: prompt template complexity, shot sampling method, and shot quantity provided in the prompt. These dimensions are assessed across four widely used LLMs (GPT-4, Mistral, Llama 3, and Flan-UL2), representing a range of larger and smaller model sizes and architectural variations to understand the adaptability of different models to in-context learning dimensions. Specifically, our study examined the following seven models:

a) *GPT-4*: Turbo (gpt-4-0125-preview) and gpt-4o-mini;
b) *Mistral*: Mixtral-8x7B-Instruct-v0.1 and Mistral-7B-Instruct-v0.2;
c) *Llama 3*: Meta-Llama-3-70B-Instruct and Meta-Llama-3-8B-Instruct;
d) *Flan-UL2*.

*Prompt template complexity*

Prompt template complexity is defined as the level of detail of the prompt, with two levels tested: basic prompt and detailed prompt. In the basic prompt, we provide basic and essential information about HPV vaccination and instruct the model to classify the stance of a tweet regarding HPV vaccination. This prompt specifies that the stance should fall into one of the three categories: "in favor," "against," or "neutral or unclear."

The detailed prompt builds on this by adding more guidance and information. In this version, we prompt the model to take on the role of "an expert content analyst," providing more specific and detailed definitions for each stance category, contextual information about HPV vaccination, and specifying the breadth of claims to consider ("statements, facts, statistics, opinions, or anecdotes"). This prompt also includes a caution to the model about the potential consequences of misclassifications, reinforcing the importance of accuracy. Examples of both prompt templates are listed in **Appendix A**.

*Shot sampling method*

The shot sampling method tested includes two approaches: random selection and stratified selection of examples (or "shots"). In random selection, a specified number of tweets is randomly drawn from the annotated training dataset, with no regard for stance balance. Stratified selection, on the other hand, involves randomly selecting a certain number of tweets representing each stance level ("in-favor," "against," and "neutral or unclear") from the annotated training dataset, to ensure a balanced representation of stance categories within the sample. This approach aims to expose the model to a more representative array of linguistic cues for each stance, which may improve classification accuracy.

*Shot quantity*

To evaluate the influence of shot quantity, we experiment with a range of shot counts, from 0 (zero-shot) up to 30 shots, with intervals of 3. In other words, the shot quantities tested include 0, 3, 6, 9, 12, 15, 18, 21, 24, 27, and 30. This systematic variation allows us to examine the trade-offs between the depth of contextual examples provided in the model's classification performance. By including zero-shot learning, we were able to assess the model's ability to infer stance without any supporting examples, providing a baseline to compare with few-shot and many-shot configurations.

*Ground-truth data*

The raw dataset used in the present study is Twitter data collected from Synthesio (www.synthesio.com), a social listening platform, using a list of search terms related to HPV vaccination from January 1, 2023, to June 28, 2023. After collecting the raw Twitter data ($N$ = 313,900), three well-trained research assistants further provided human annotations of a random sample ($n$ = 1,050) for stance identification ("in-favor," "against," or "neutral or unclear") toward HPV vaccination as the ground-truth dataset for the study. We acknowledge that some tweet content may contain ambivalent and contradictory stances and information. To reduce stance ambivalence, we excluded tweets for which there was disagreement among the three research assistants on stance annotations. By doing this, the total volume of this final ground-truth dataset is 756, including 367 tweets annotated as “in-favor,” 327 tweets annotated as “against,” and 62 tweets annotated as “neutral or unclear” by the research assistants. Three annotators had unanimous agreement on these annotations.

*Procedures*

We investigated the performances of various instruction fine-tuned LLMs, in classifying the stance of the tweets toward HPV vaccination, using zero-shot and few-shot configurations, with varying shot quantities. The tested LLMs, GPT-4 (including GPT-4 Turbo and GPT-4o-mini), Mistral (including Mixtral-8x7B-Instruct-v0.1 and Mistral-7B-Instruct-v0.2), Llama 3 (including Meta-Llama-3-70B-Instruct and Meta-Llama-3-8B-Instruct), and Flan-UL2, were prompted to classify the stance based on prompts specified in the previous section (3.1). The performance of the model within a given in-context learning scenario was evaluated using macro F1 score–a metric that treats the precision of the model on all three stance categories equally, regardless of the relative abundance of posts of that stance in the dataset. Of all possible metrics, the macro F1 score best aligns with the aims of this study as it weighs all three categories equally and is uninfluenced by the relative ampleness of examples of each category in the evaluation dataset (Opitz, 2022).

In the zero-shot setting, LLMs were requested to classify any given tweet's stance without any examples provided, relying solely on the prompt instructions of the task, which included either a basic or detailed template as outlined in Section 3.1. In contrast, the few-shot settings provided additional contextual guidance through a selected number of in-context examples, accompanying each task prompt. The presence of shots in the prompt allowed the LLMs to infer patterns in stance classification by observing in-context examples labeled as "in favor," "against," or "neutral/unclear" regarding HPV vaccination. To maintain data integrity and balance, we split the final annotated dataset into training and test sets in a stratified 50-50 manner. Stratified sampling ensured that stance categories were evenly represented across both sets, thus preserving stance balance for reliable performance evaluation. Tweets in the training set were used as few-shot examples within the prompts, while tweets in the test set were used to evaluate stance classification performance for each LLM configuration.

To systematically evaluate the three dimensions in the experimental design—prompt template complexity, shot sampling method, and shot quantity—we created a comprehensive prompt dataset by embedding each test tweet within multiple prompt configurations. The total number of prompt configurations was determined by the combination of the following levels: prompt template complexity (basic v.s. detailed), shot sampling methods (random v.s. stratified), and shot quantity (ranging from 0 to 30 shots in intervals of 3). For each tweet in the test set, we generated 40 few-shot prompts and two zero-shot prompts, resulting in a dataset of 15,876 unique prompts. Specifically, each few-shot prompt for a test tweet was created by sampling examples from the training set independently. This approach ensured that every few-shot prompt provided unique combinations of contextual examples, maintaining diversity and reducing potential overlap that could lead to repetitive cues in classification. **Figure 1** shows the high-level process for creating prompts for any test tweet.

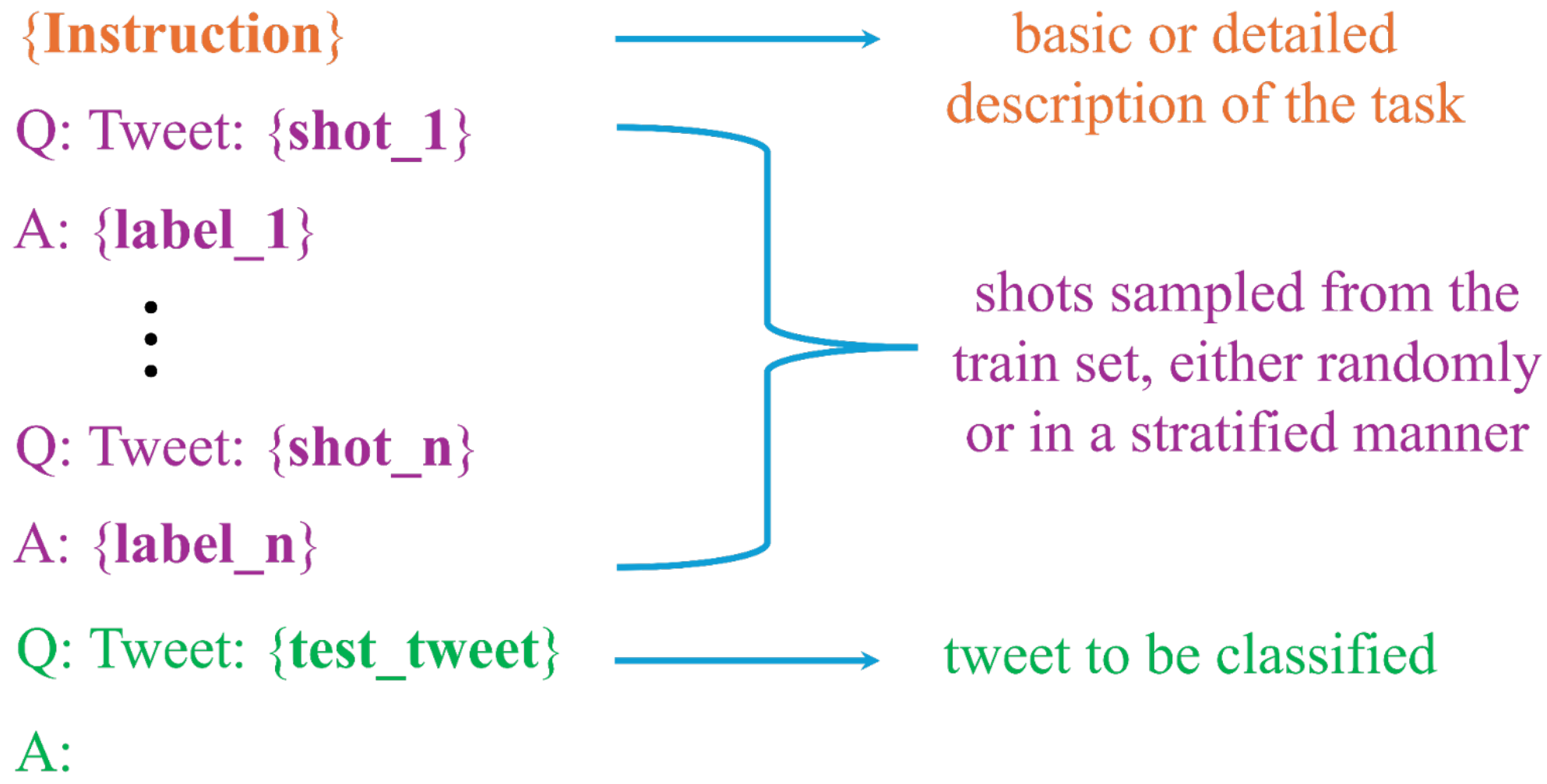

***Figure 1.*** Overview of the prompt creation procedure.

*Inference*

Given that GPT-4 models are closed-source, we used OpenAI's APIs to send our inference requests directly to their servers. For all the other open-source models—Flan-UL2, Mistral, and Llama 3—we obtained their pre-trained weights from Hugging Face's Transformers library and conducted inference locally on the authors' institute's server. This server was equipped with two NVIDIA RTX 6000 Ada Generation GPUs, each with 48GB of memory, enabling efficient handling of large-scale computations.

For each LLM, we tokenized prompts using the respective tokenizer classes from the Hugging Face's Transformers library, to ensure compatibility with each LLM's specific architecture. Since these models support different maximum context lengths, we excluded any prompts that exceeded a model's input capacity. The maximal context lengths are as follows: 128,000 tokens for GPT-4 models, 8,192 tokens for Llama 3 models, 32,768 tokens for Mistral models, and 2,048 tokens for Flan-UL2.

Given that we aimed to optimize model focus and consistency for the classification task, we set the temperature parameter to zero or close to zero to minimize the randomness of the output. Specifically, we set the temperature to 0 for GPT-4 models, and 1e-5 for Llama 3, Mistral, and Flan-UL models. Additionally, we standardized certain parameters across these LLMs: 1) the batch size was set to 1 for inference processing speed, and 2) the maximum output length was capped at 200 tokens for all the models to control output verbosity. For memory efficiency, Mistral and Llama 3 models were loaded in 4 bits, reducing GPU memory consumption while maintaining computational efficiency.

*Post-processing the LLM outputs*

Since the LLM outputs do not always align exactly with the predefined stance labels, we employed a post-processing strategy using a pattern-matching tool to reliably extract the predicted stance label from each model output. Specifically, for each raw output, if the completion explicitly begins with or includes only one stance label (i.e., "in favor," "against," and "neutral or unclear"), we treated the label as the model's prediction. Otherwise, in cases where multiple stance labels appeared in the response or where the response was ambiguous, we manually inspected the raw output and assigned the correct label. This human-in-the-loop approach ensured that the majority of completions were correctly categorized, allowing us to handle exceptions effectively and maintain high accuracy in the task of stance detection.

***3.2. Fine-tuning methods and model selection***

In the fine-tuning modeling, we employed a popular Parameter-Efficient Fine-Tuning (PEFT) technique called Low-Rank Adaptation (LoRA), which decomposes matrices within certain subunits of LLMs, reducing the number of trainable parameters in the model and the computational resources needed to fine-tune it (Hu et al., 2021). LLMs fine-tuned with LoRA perform as well as or better than models fine-tuned classically without using PEFT techniques (Hu et al., 2021). Three models were chosen for fine-tuning with LoRA. These three models were chosen from amongst those used for in-context learning-based stance detection to serve as a representative sample of the models. The three models are the most performant models of all families of LLMs evaluated during the in-context learning-based stance analysis, with the exclusion of the GPT-4-Turbo, which is an OpenAI proprietary model and, therefore, cannot be fine-tuned. Hence, the fine-tuned models were Flan-UL2, Meta-Llama-3-70B-Instruct, and Mixtral-8x7B-Instruct-v0.1.

## 4. Results

### *4.1. In-context learning model performance*

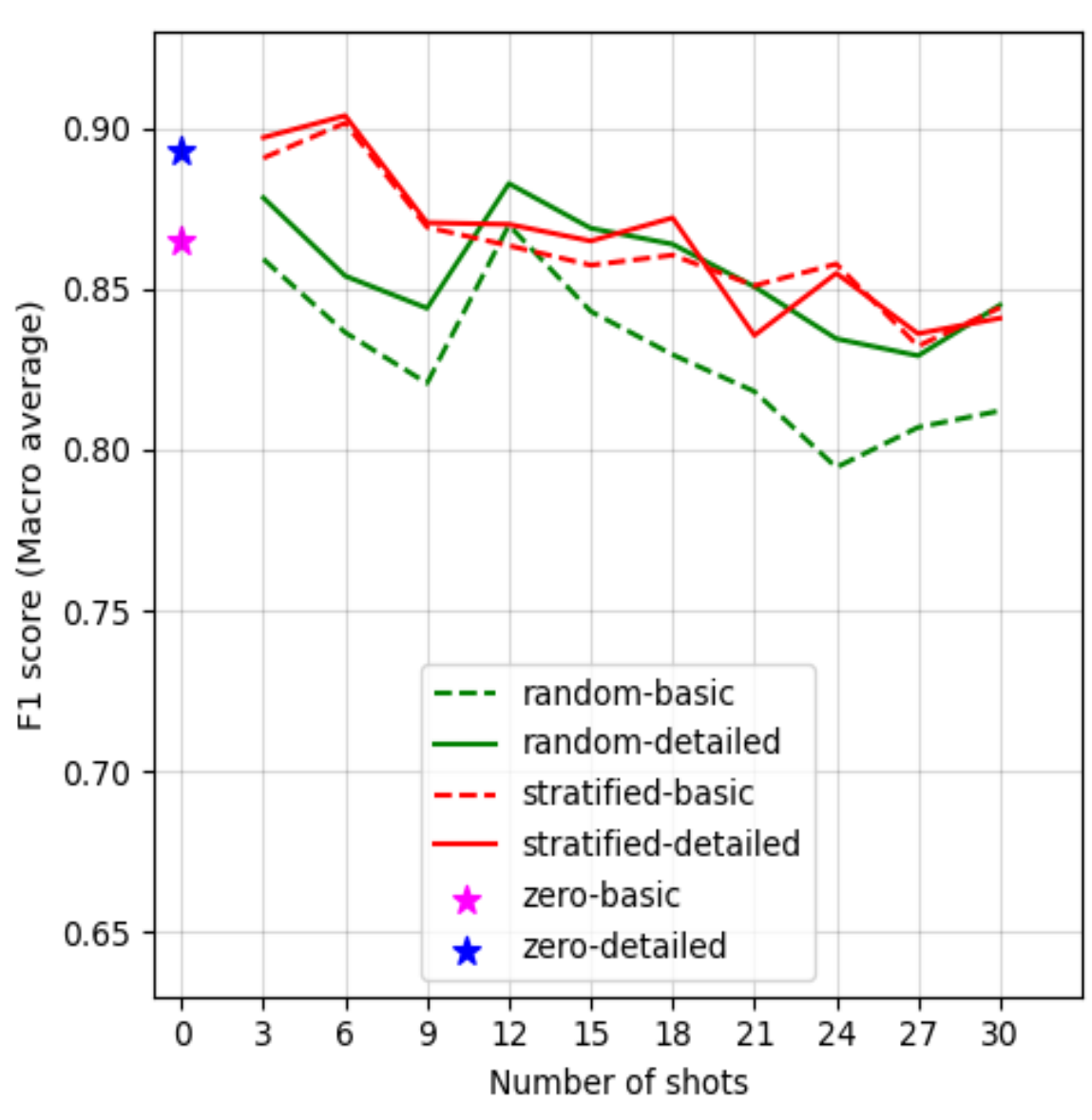

**(A).** F1 scores of *GPT-4 Turbo* of in-context learning.

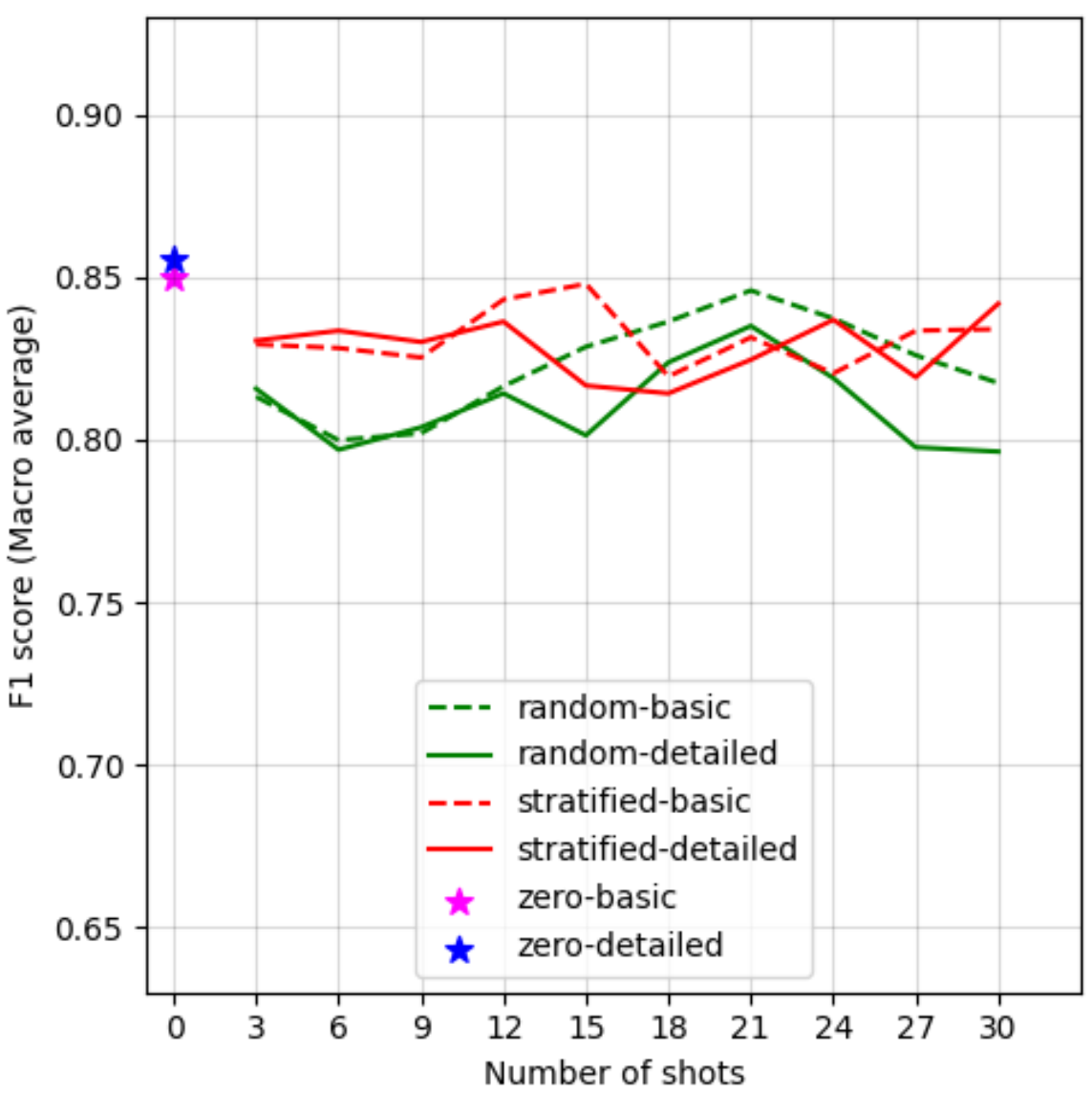

**(B).** F1 scores of *GPT-4o-mini* of in-context learning.

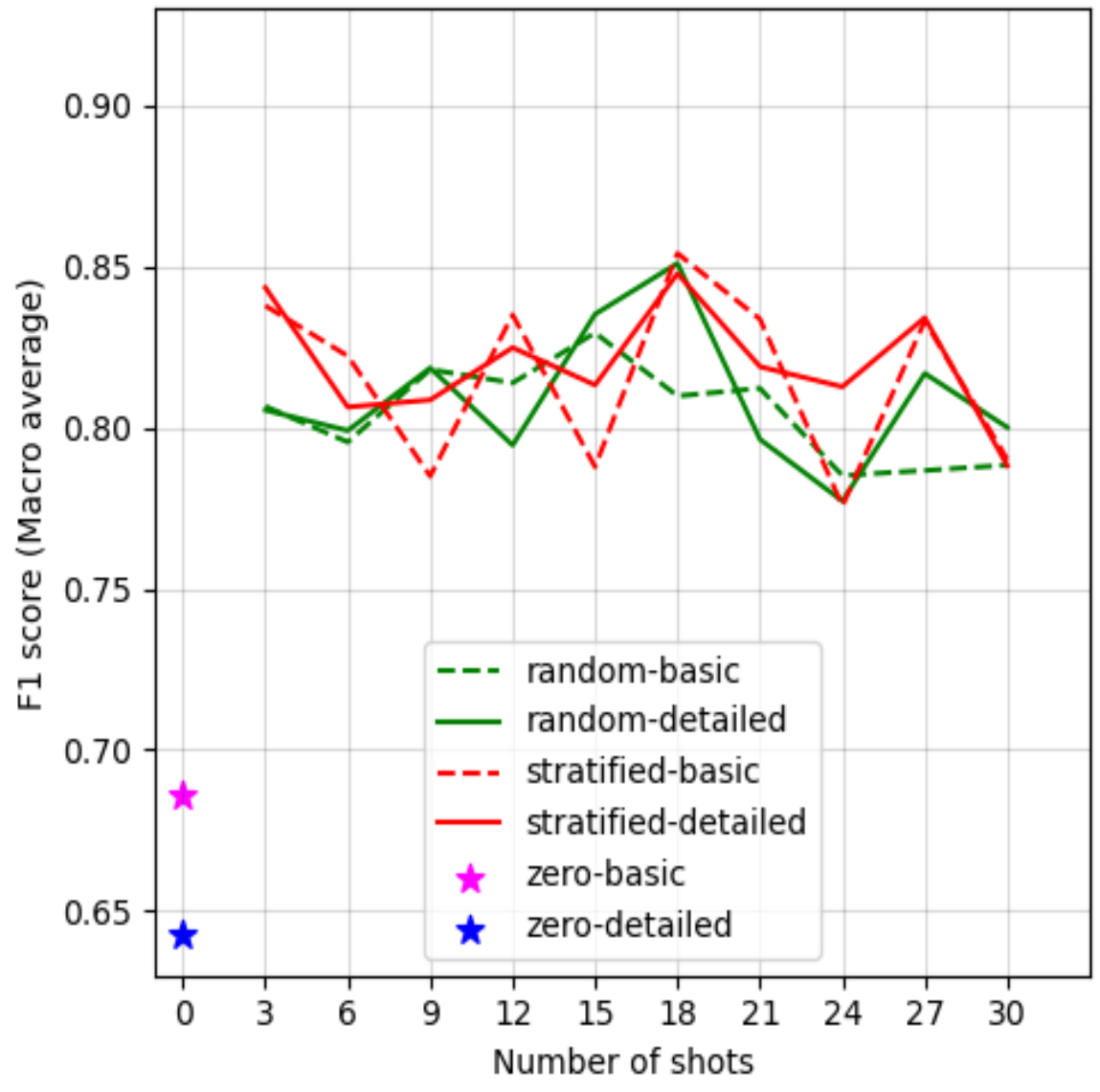

**(C).** F1 scores of *Mixtral-8x7B-Instruct* of in-context learning.

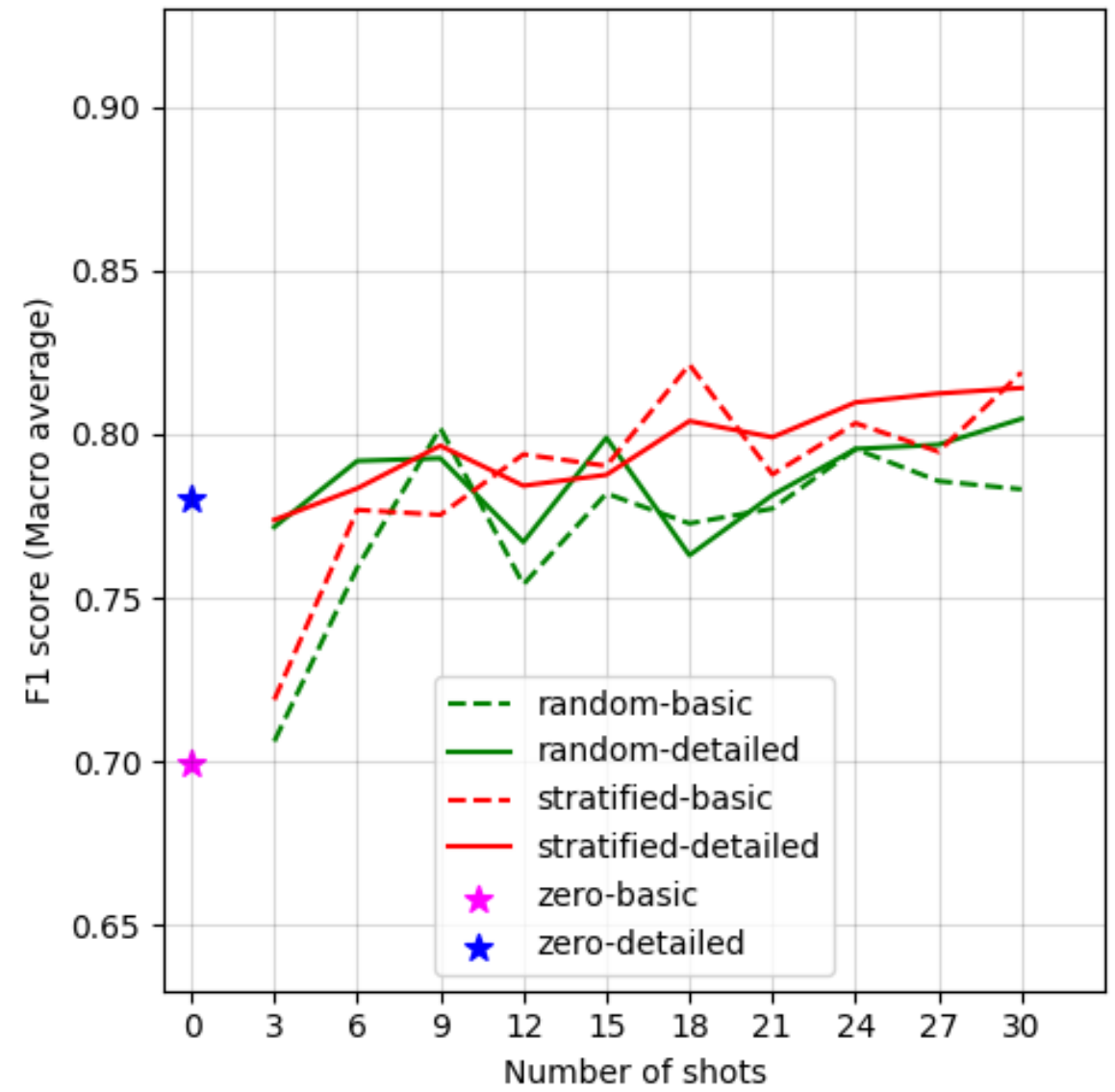

**(D).** F1 scores of *Mistral-7B-Instruct* of in-context learning.

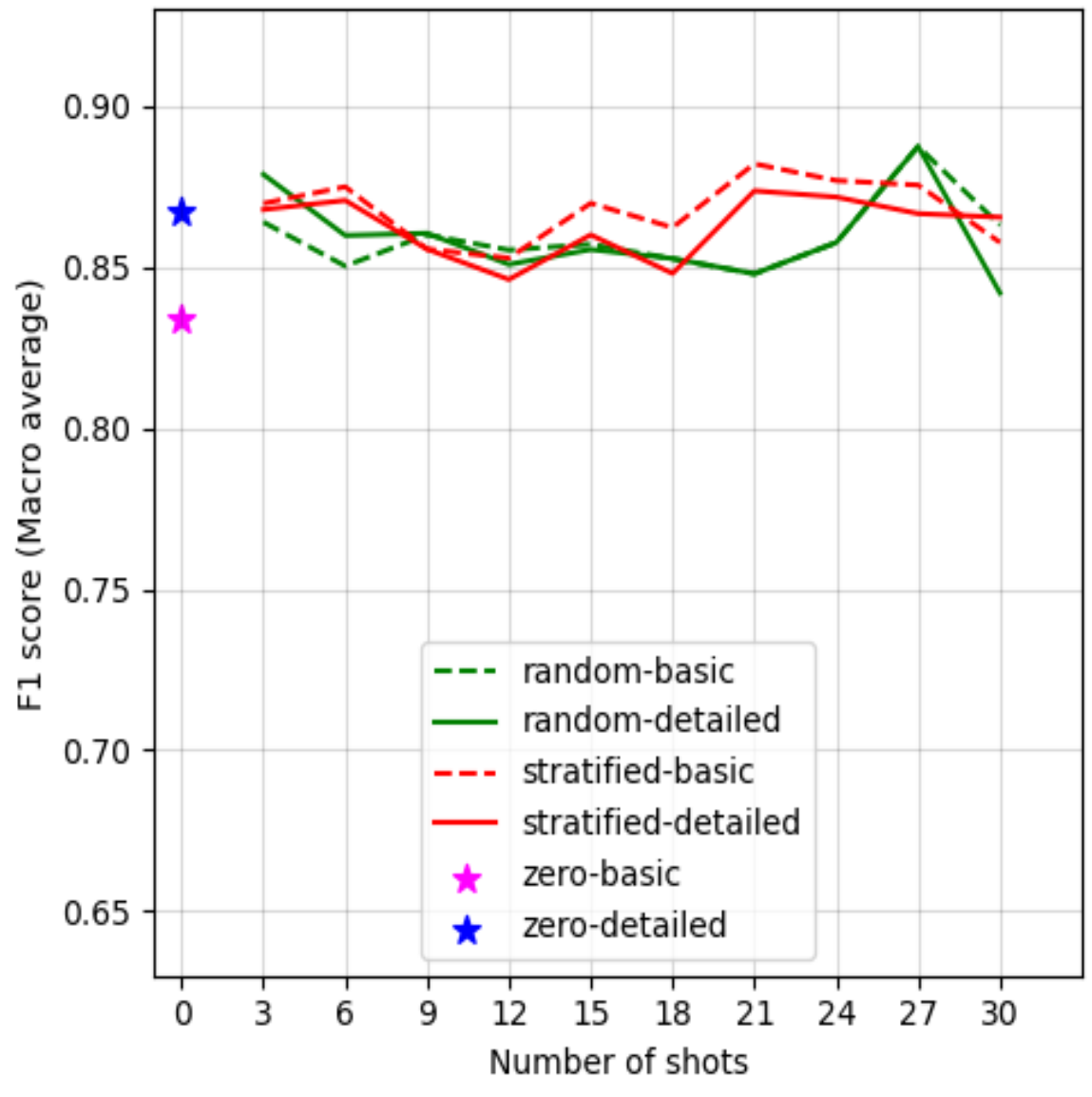

**(E).** F1 scores of *Llama-3-70B-Instruct* of in-context learning.

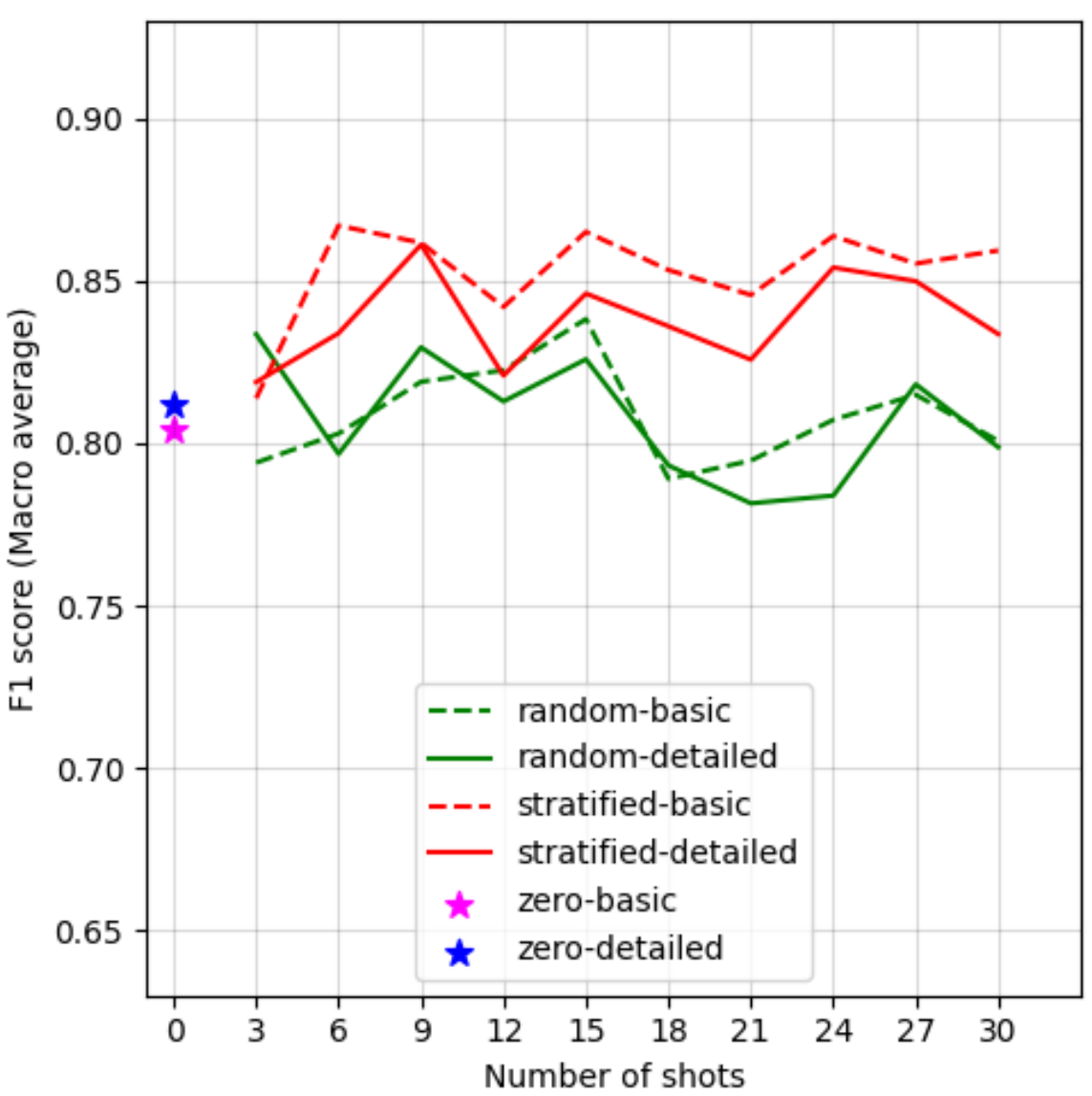

**(F).** F1 scores of *Llama3-8B-Instruct* of in-context learning.

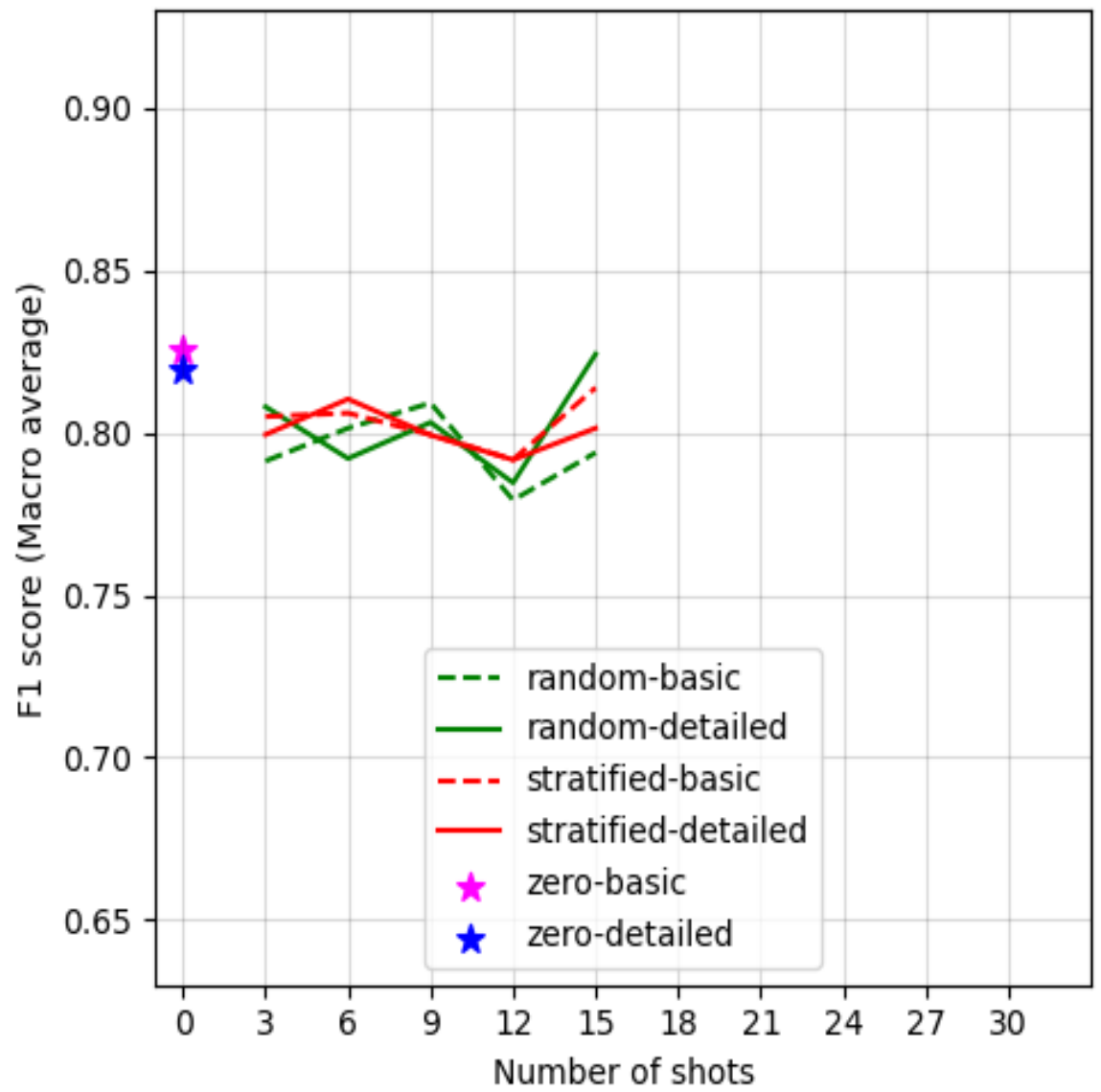

**(G).** F1 scores of *Flan-UL2* of in-context learning.

***Figure 2.*** Performance of LLMs across varying experimental dimensions of in-context learning.

**Figures 2(A)** to **Figure 2(G)** display the F1 scores for each LLM, plotted against the number of shots and prompt conditions (prompt template complexity and shot sampling methods) in in-context learning. Our experimental test results show varied performance trends across different models and conditions. Overall, GPT-4 Turbo presents the highest performance among all LLMs. Notably, most models achieve acceptable F1 scores in the zero-shot learning conditions, though Mixtral-8x7B-Instruct-v0.1 falls short, with F1 scores below 0.7. However, performance improves across models by changing the number of shots, as well as varying the shot selection methods, though the extent of this improvement varies by model. In general, larger model variants perform better under the detailed-prompt condition (represented by solid lines in Figure 2), often outperforming the basic-prompt condition (dashed lines in Figure 2). In addition, the stratified sampling condition often yields higher F1 scores than random sampling, with the performance gap becoming more pronounced in smaller model variants.

Specifically, a unique trend appears with GPT-4 Turbo in **Figure 2(A)**, where F1 scores generally decline as the number of shots increases, except in the stratified sampling condition. Here, performance peaks at 6 shots, achieving the highest F1 score of 0.90 under the detailed prompt. In other words, this 6-shot, stratified sampling with a detailed prompt performs as the optimal configuration for GPT-4 Turbo. Interestingly, increasing the number of shots does not improve model performance for GPT-4o-mini in **Figure 2(B)**, with zero-shot conditional performing the best (F1 scores = 0.85 and 0.86 in basic and detailed prompts).

Conversely, zero-shot performance for Mixtral-8x7B-Instruct-v0.1 is comparatively lower, with F1 scores of 0.64 for a detailed prompt, and 0.69 for a basic prompt, as shown in **Figure 2(C)**. However, by simply adding three few-shot examples, F1 scores substantially improved by 0.20 for detailed prompts, and by 0.12 for basic prompts. This result suggests that Mixtral-8x7B-Instruct-v0.1 is highly responsive to the presence of few-shot examples. We also tested its smaller variant, Mistral-7B-Instruct, as shown in **Figure 2(D)**, revealing that Mistral-7B-Instruct is particularly sensitive to prompt template complexity. This sensitivity results in a performance gap between basic and detailed prompts, especially in the zero- and 3-shot conditions.

In the case of Llama-3-70B-Instruct, as shown in **Figure 2(E)**, it is interesting to note that the overall performance pattern follows a "U-shaped curve," peaking at the 27-shot condition with stratified sampling. Few-shot configurations with fewer than 12 or more than 21 examples outperform those in the 15–18 range, along with a slight dip at 30 shots. Nevertheless, the zero-shot condition with a basic prompt underperforms other conditions in Llama-3-70B-Instruct.

Additionally, **Figure 2(F)** shows the results of its smaller variant, Llama3-8B-Instruct, with a pronounced sensitivity to shot sampling methods: stratified sampling conditions often outperform random sampling. This may suggest that stratified examples in the prompt help Llama3-8B-Instruct acquire more balanced contextual knowledge across stance levels.

Due to Flan-UL2's context length limitation of 2,048 tokens, we do not show the F1 scores for prompt configurations with more than 15 shots, as fewer than 100 tweets meet this context length criterion with higher shot counts. Nonetheless, we observe an uptick in performance for the detailed prompt at 15 shots, approaching zero-shot performance, though it still lags behind GPT-4 or Llama-3-70B-Instruct.

Notably, our experimental procedure revealed the occurrence of ill-formatted text outputs such as "missing initial labels," "irrelevant stances," and "dual stances" (detailed categories provided in **Appendix D**). The ill-formatted output issue can impact the performance of the classification task (de Wynter et al., 2023). Specifically, we noticed that zero-shot prompts led to an increase in "missing initial label" output, where completions often included unnecessary reasoning or failed to start with a label. Fortunately, we were able to resolve all the cases by integrating a human-in-the-loop approach described in the Methods section.

### ***4.2. Fine-tuning model comparison***

After fine-tuning three models (i.e., Mixtral-8x7B-Instruct-v0.1, Meta-Llama-3-70B-Instruct, and Flan-UL2) with LoRa, the Flan-UL2 model was found to be the most performant based on macro F1 score, narrowly outperforming the Mistral model and surpassing the Meta-Llama-3-70B-Instruct model by a sizeable margin. The complete performance metrics of all three models fine-tuned in this study are shown in **Table 1**. The superior performance of the Flan-UL2 model may seem initially surprising as it performs markedly worse than the other two models on popular LLM benchmarks (Dubey et al., 2024; Jiang et al., 2024; Tay et al., 2023). However, given that more of Flan-UL2's parameters are accessible and amenable to LoRA fine-tuning relative to its total architecture than in Mixtral-8x7B-Instruct-v0.1 and Meta-Llama-3-70B-Instruct, it is foreseeable that it will perform more prodigiously than the other models as it was changed to a greater extent by the fine-tuning process. Additionally, both the Flan-UL2 and Mixtral-8x7B-Instruct-v0.1 models are significantly smaller than the Meta-Llama-3-70B-Instruct model in terms of parameters. The Flan-UL2 and Mixtral-8x7B-Instruct-v0.1 models have twenty and seven billion parameters, respectively, while the Meta-Llama-3-70B-Instruct model has seventy billion (Dubey et al., 2024; Jiang et al., 2024; Tay et al., 2023). We speculate that Flan-UL2's classification capabilities and Mixtral-8x7B-Instruct-v0.1's comparable classification prowess, which is only marginally smaller, are better than Meta-Llama-3-70B-Instruct’s because of their smaller sizes, which allow them to respond more readily to and retain the benefits of fine-tuning, especially on a smaller dataset with a stance category imbalance (Kalajdzievski, 2024; Luo et al., 2024).

**Table 1.**
*Complete Performance Metrics of the Three Models Fine-Tuned with LoRa*

| Model | Micro F1 Score | Macro F1 Score | Weighted F1 Score |
|---|---:|---:|---:|
| Mixtral-8x7B-Instruct-v0.1 | 0.8919 | 0.8056 | 0.9025 |
| Flan-UL2 | 0.8829 | **0.8126** | 0.8909 |
| Meta-Llama-3-70B-Instruct | 0.8649 | 0.7691 | 0.8787 |

*Note*. F1 scores are presented in four decimal places to highlight specific performance differences among the three models.

Overall, each model’s performance during inference on the evaluation dataset was generally poorer than that of their untuned counterparts employing in-context learning across almost all experimental dimensions. Moreover, even in the instances in which a model performed better than its in-context learning counterpart—such as Flan-UL2 when compared to its counterpart utilizing few-shot learning—the advantage of the fine-tuned model was only slight and not consistent across differing numbers of in-context examples. The fact that fine-tuning in our study provided little improvement in model performance compared to in-context learning is an appropriate result given that modern models, such as the Meta-Llama-3-70B-Instruct and Mixtral-8x7B-Instruct-v0.1 models, are trained on prompts with in-context examples, making them intrinsically more receptive to and performant with in-context learning methodologies (Dubey et al., 2024; Jiang et al., 2024). Additionally, previous studies have found that on smaller datasets, LLMs that are fine-tuned generally underperform relative to LLMs employing in-context learning, further supporting the validity of the results of the comparison (Bertsch et al., 2024).

## 5. Discussion and implications

Our study contributes to the application of computational social science by confirming and recommending the feasibility of integrating LLMs into social science research. Additionally, our detailed evaluation of the performances of different practices with empirical evidence provides critical insights into how to use LLMs effectively. This study aims to elucidate how different practices may yield different results and performances, and which dimensions and conditions researchers should consider when integrating LLMs for specific research purposes.

We consider three dimensions of in-context learning in the study: prompt template complexity, shot sampling method, and shot quantity, as these are crucial in the information provided in in-context learning. We compared the performances of four types of LLMs and their variants, and found that their performance patterns and sensitivity varied across conditions and models. Take the most widely used LLM, GPT4-Turbo, as an example, we find that the most accurate and efficient approach for a stance annotation task is to provide six shots of examples using a stratified sampling approach (two examples for each stance level), and a detailed, contextual prompt to prompt GPT-4 Turbo to work as an expert analyst.

Moreover, the overall decreasing pattern of the performance metrics for GPT-4 Turbo suggests the phenomenon of *cognitive overload* may also occur in in-context learning (Upadhayay et al., 2024). In other words, providing too much information may not yield better decisions, which could occur to both human beings and LLMs. It is likely that providing more shots of examples as references in the prompt for classifying stance creates an extra burden for information processing, making it less confident to identify the stance of the assigned text. We acknowledge that this phenomenon only occurs in GPT-4 Turbo and not in other LLMs; therefore, further investigations are encouraged before making any generic conclusions.

We also take fine-tuning methods into consideration, and conducted model comparison with in-context learning performance. It is worth noting that our results show that fine-tuning models do not consistently outperform their untuned counterparts employing in-context learning. The limited improvement in model performance with fine-tuning compared to in-context learning in our study aligns with expectations, as modern models like Meta-Llama-3-70B-Instruct and Mixtral-8x7B-Instruct-v0.1 are pre-trained to work effectively with prompts containing in-context examples, making them inherently responsive to in-context learning methods (Dubey et al., 2024; Jiang et al., 2024). Moreover, prior research has shown that on smaller datasets, fine-tuned LLMs often underperform relative to those employing in-context learning, further validating our comparison results (Bertsch et al., 2024).

In spite of the fact that the fine-tuned models do not regularly best the models employing in-context learning, they may still be preferable to in-context learning models in certain situations because the computational resources needed for inference with a fine-tuned model are significantly less than those required for models utilizing in-context learning, meaning that inference is often faster and more scalable with fine-tuned models (Mosbach et al., 2023). This key benefit is particularly crucial when computational resources are limited, the processing task is enormous, or reducing the length of model inference is paramount.

Future designs may want to scrutinize more nuanced aspects of prompt design and in-context learning when applying LLMs. For instance, our current experimental design includes the factor of prompt template complexity, in which we provided two conditions: basic and detailed prompts. The overall results show that detailed-prompting conditions yield better performance compared to the basic-prompting conditions. In the detailed version of the prompt, we provided four additional details: 1) prompting the model to work as "an expert content analyst," 2) including detailed definitions of the three levels of stance (i.e., "in-favor," "against," and "neutral or unclear") with specific contextual information about HPV vaccination, 3) specified the breadth of claims to consider in stance detection ("statements, facts, statistics, opinions, or anecdotes"), and 4) alerting the model of costly consequences if misclassified. Since multiple elements were altered between the detailed-prompting condition and the basic-promoting condition, it is unclear which aspect of the message design helped the most in terms of improving the performance of the LLMs. Future research should test more precise research designs to understand which prompt design components were consequential for LLMs' classification tasks.

Another issue that future research should address is that our ground-truth dataset is restricted to the subset of unambiguous tweets. It is uncertain whether including more ambiguous tweets in the shot example pool would yield different results. It is possible that the overall performance would decline as LLMs get confused by the ambiguous content provided in the prompts. Conversely, it could also be possible that the overall performance improves as LLMs gain more information and expand their knowledge span to classify ambiguous tweets based on seeing more edge cases. Future research should also take this unanimity-ambiguity dimension further into account.

Last, the theme and its distribution in the data, especially those concerning vaccine hesitancy and misinformation, may also affect the performance of LLMs. Researchers have found that concerns about vaccine effectiveness and safety have been prevalent around online discourses regarding vaccination (Dhaliwal & Mannion, 2020; Di Domenico et al., 2022; Massey et al., 2020). We were also able to annotate the particular themes of the vaccine discourses in our ground-truth dataset, including reproductive health concerns and misinformation connecting HPV vaccination to death and cancer. However, we did not take the theme dimensions into account to maintain a relatively simple and straightforward experimental design. Future research should consider the specific contextual themes of the datasets and include them as dimensions in the stratification of the shot selection.

We acknowledge that the present study serves as a case study, for demonstrating the use of LLMs, particularly in-context learning, in computational social science research. Our LLM task focused solely on stance classification. Other more advanced and creative tasks in different contexts, other than classification using Twitter data, may need to consider more nuanced dimensions of prompt engineering. Last but not least, we should keep in mind the importance of human-in-loop processes to avoid biases from LLMs and AI techniques in general. Although our experimental test in the context of HPV vaccination design provides crucial directions for computational social science, it also has implications for health communication and political communication research. The ability to classify and analyze content on a public health issue such as HPV vaccination, at scale, could be used to monitor the spread of low-credibility information, develop mitigation and intervention strategies for this problematic content, and support public health workers seeking to counter vaccine hesitancy. For those focused on political outcomes, this same classifier could prove useful for understanding the partisanship of vaccine skepticism and misinformation, the politicized policy debate surrounding HPV vaccines, and the role of political elites in amplifying misinformation.

## 6. Conclusion

To sum up, in the present study, we conduct an application experiment for a stance classification task using both in-context learning and fine-tuning approach, comparing various LLMs and their variants, including GPT-4 (Turbo and GPT-4o-mini), Mistral (Mixtral-8x7B-Instruct-v0.1 and Mistral-7B-Instruct-v0.2), Llama 3 (Meta-Llama-3-70B-Instruct and Meta-Llama-3-8B-Instruct), and Flan-UL2, in the context of HPV vaccine tweets on social media. Our experimental design for in-context learning focused on three dimensions: prompt template complexity (basic v.s. detailed), shot sampling method (random v.s. stratified), and shot quantity provided in the prompt (0 to 30, with intervals of 3). We also took the more conventional fine-tuning methods into consideration, examining whether providing in-context knowledge would yield a higher performance across models. The contextual dataset used in the study, consisting of social media data discussing HPV vaccination on Twitter, pre-labeled with three levels of stance identification (i.e., “in-favor,” “against,” and “neutral or unclear”) by well-trained human annotators.

Echoing Ziems et al. (2024), we recommend integrating LLMs for computational social science research, particularly applying GPT-4 Turbo due to its higher overall performance metrics in the present study. Specifically, our findings show that more shots of examples do not guarantee better performance for stance classification, particularly for the most widely applied LLM, GPT-4 Turbo. Moreover, the performance sensitivity of these models varies by prompt template complexity and shot selection methods. The best practice observed in our study is to provide six shots of examples using a stratified sampling approach (two examples for each stance level) and a detailed, contextual prompt to prompt GPT-4 Turbo to work as an expert analyst for the stance classification task. Other LLMs, such as Mistral, Llama 3, and Flan-UL2, exhibited different patterns of performance across the three main dimensions of in-context learning in the experiment.

## 8. Appendices

***Appendix A.*** *Basic and detailed prompt templates used in the LLM experimental study*

1. **Basic prompt:**

What is the stance of the tweet below with respect to vaccination against human papilloma virus, often abbreviated as HPV? If we can infer from the tweet that the Tweeter supports human papilloma virus (HPV) vaccination, please label it as "in favor." If we can infer from the tweet that Tweeter is against human papilloma virus (HPV) vaccination, please label it as "against." If we can infer from the tweet that the Tweeter has a neutral stance towards human papilloma virus (HPV) vaccination, please label it as "neutral or unclear." If there is no indication in the tweet to reveal the stance of the Tweeter or towards human papilloma virus (HPV) vaccination, please also label it as "neutral or unclear." Please use only one of the following three categories labels to classify its stance: "in favor," "against," or "neutral or unclear." Here are some examples of tweets that are "in favor," "against," or "neutral or unclear" to provide you guidance. Make a strong effort to classify the last tweet correctly.

2. **Detailed prompt:**

You are an expert content analyst with experience classifying the stance of text. What is the stance of the tweet below with respect to vaccination against human papilloma virus, often abbreviated as HPV? If we can infer from the tweet that the Tweeter supports human papilloma virus (HPV) vaccination, please label it as "in favor." By "in-favor," we mean providing supportive statements, facts, statistics, opinions, or anecdotes that (a) endorse vaccination in general, (b) mention the health benefits of vaccination, or (c) emphasize its effectiveness in preventing infection from HPV, averting precancerous lesions, and reducing the risk of cancer and death. If we can infer from the tweet that Tweeter is against human papilloma virus (HPV) vaccination, please label it as “against.” By “against,” we mean providing skeptical or inaccurate statements, facts, statistics, opinions, or anecdotes that (a) oppose vaccination in general, (b) question the health benefits of vaccination, or (c) link vaccination to reproductive health and pregnancy risks, the increased possibility of developing cancer, and greater likelihood of death or serious medical complications. If we can infer from the tweet that the Tweeter has a neutral stance towards human papilloma virus (HPV) vaccination, please label it as “neutral or unclear.” By “neutral or unclear” we mean balancing benefits of HPV vaccination against potential risk of vaccination without a clear tilt in favor or against vaccination. If there is no indication in the tweet to reveal the stance of the Tweeter or towards human papilloma virus (HPV) vaccination, please also label it as “neutral or unclear.” Please use only one of the following three categories labels to classify its stance: “in favor,” “against,” or “neutral or unclear.” Here are some examples of tweets that are “in favor,” “against,” or “neutral or unclear” to provide you guidance. Make a strong effort to classify the last tweet correctly, as misclassifications may have costly consequences.

***Appendix B.*** *Search terms for raw data collection via Synthesio*

(hpv vaccine) OR (hpv vaccination) OR (hpv vaccinate) OR (hpv vax) OR (hpv vaxxed) OR (hpv jab) OR (hpv jabbed) OR (hpv shot) OR ("human papillomavirus" vaccine) OR ("human papillomavirus" vaccination) OR ("human papillomavirus" vaccinate) OR ("human papillomavirus" vax) OR ("human papillomavirus" vaxxed) OR ("human papillomavirus" shot) OR ("human papillomavirus" jab) OR ("human papillomavirus" jabbed) OR gardasil OR cervarix.

***Appendix C.*** *In-context learning LLM performance metrics*

**Table C1.**
*F1 scores of GPT-4 models*

| | | GPT4 Turbo | | GPT-4o-mini | |
|---|---|---|---|---|---|
| Shot # | Prompt Template | Weighted F1 Score | Macro F1 Score | Weighted F1 Score | Macro F1 Score |
| Random sampling | | | | | |
| 3 | Basic | 0.93 | 0.86 | 0.91 | 0.81 |
| 3 | Detailed | 0.94 | 0.88 | 0.91 | 0.82 |
| 6 | Basic | 0.92 | 0.84 | 0.90 | 0.80 |
| 6 | Detailed | 0.93 | 0.85 | 0.90 | 0.80 |
| 9 | Basic | 0.91 | 0.82 | 0.90 | 0.80 |
| 9 | Detailed | 0.93 | 0.84 | 0.90 | 0.80 |
| 12 | Basic | 0.94 | 0.87 | 0.91 | 0.82 |
| 12 | Detailed | 0.95 | 0.88 | 0.91 | 0.81 |
| 15 | Basic | 0.92 | 0.84 | 0.91 | 0.83 |
| 15 | Detailed | 0.94 | 0.87 | 0.90 | 0.80 |
| 18 | Basic | 0.91 | 0.83 | 0.91 | 0.84 |
| 18 | Detailed | 0.93 | 0.86 | 0.91 | 0.82 |
| 21 | Basic | 0.91 | 0.82 | 0.93 | 0.85 |
| 21 | Detailed | 0.93 | 0.85 | 0.92 | 0.83 |
| 24 | Basic | 0.89 | 0.79 | 0.92 | 0.84 |
| 24 | Detailed | 0.92 | 0.83 | 0.91 | 0.82 |
| 27 | Basic | 0.90 | 0.81 | 0.91 | 0.83 |
| 27 | Detailed | 0.92 | 0.83 | 0.90 | 0.80 |
| 30 | Basic | 0.91 | 0.81 | 0.91 | 0.82 |
| 30 | Detailed | 0.93 | 0.84 | 0.90 | 0.80 |
| Stratified sampling | | | | | |
| 3 | Basic | 0.95 | 0.89 | 0.92 | 0.83 |
| 3 | Detailed | 0.95 | 0.90 | 0.91 | 0.83 |
| 6 | Basic | 0.95 | 0.90 | 0.91 | 0.83 |
| 6 | Detailed | 0.96 | 0.90 | 0.91 | 0.83 |
| 9 | Basic | 0.94 | 0.87 | 0.91 | 0.83 |
| 9 | Detailed | 0.94 | 0.87 | 0.91 | 0.83 |
| 12 | Basic | 0.94 | 0.86 | 0.92 | 0.84 |
| 12 | Detailed | 0.94 | 0.87 | 0.91 | 0.84 |
| 15 | Basic | 0.93 | 0.86 | 0.93 | 0.85 |
| 15 | Detailed | 0.93 | 0.86 | 0.90 | 0.82 |
| 18 | Basic | 0.93 | 0.86 | 0.91 | 0.82 |
| 18 | Detailed | 0.94 | 0.87 | 0.90 | 0.81 |
| 21 | Basic | 0.93 | 0.85 | 0.91 | 0.83 |
| 21 | Detailed | 0.92 | 0.84 | 0.91 | 0.82 |
| 24 | Basic | 0.93 | 0.86 | 0.91 | 0.82 |
| 24 | Detailed | 0.93 | 0.85 | 0.91 | 0.84 |
| 27 | Basic | 0.92 | 0.83 | 0.91 | 0.83 |
| 27 | Detailed | 0.92 | 0.84 | 0.90 | 0.82 |
| 30 | Basic | 0.93 | 0.84 | 0.92 | 0.83 |
| 30 | Detailed | 0.93 | 0.84 | 0.92 | 0.84 |
| Zero-shot | | | | | |
| 0 | Basic | 0.94 | 0.86 | 0.93 | 0.85 |
| 0 | Detailed | 0.96 | 0.89 | 0.93 | 0.86 |

**Table C2.**
*F1 scores of Mistral models*

| Shot # | Prompt Template | Mixtral-8x7B-Instruct Weighted F1 Score | Mixtral-8x7B-Instruct Macro F1 Score | Mistral-7B-Instruct Weighted F1 Score | Mistral-7B-Instruct Macro F1 Score |
|---|---|---|---|---|---|
| Random sampling | | | | | |
| 3 | Basic | 0.91 | 0.81 | 0.82 | 0.71 |
| 3 | Detailed | 0.91 | 0.81 | 0.88 | 0.77 |
| 6 | Basic | 0.90 | 0.80 | 0.87 | 0.76 |
| 6 | Detailed | 0.91 | 0.80 | 0.89 | 0.79 |
| 9 | Basic | 0.92 | 0.82 | 0.90 | 0.80 |
| 9 | Detailed | 0.92 | 0.82 | 0.90 | 0.79 |
| 12 | Basic | 0.92 | 0.81 | 0.87 | 0.75 |
| 12 | Detailed | 0.91 | 0.79 | 0.88 | 0.77 |
| 15 | Basic | 0.93 | 0.83 | 0.88 | 0.78 |
| 15 | Detailed | 0.93 | 0.84 | 0.89 | 0.80 |
| 18 | Basic | 0.92 | 0.81 | 0.88 | 0.77 |
| 18 | Detailed | 0.93 | 0.85 | 0.88 | 0.76 |
| 21 | Basic | 0.92 | 0.81 | 0.88 | 0.78 |
| 21 | Detailed | 0.92 | 0.80 | 0.89 | 0.78 |
| 24 | Basic | 0.90 | 0.79 | 0.90 | 0.80 |
| 24 | Detailed | 0.90 | 0.78 | 0.90 | 0.80 |
| 27 | Basic | 0.91 | 0.79 | 0.89 | 0.79 |
| 27 | Detailed | 0.92 | 0.82 | 0.90 | 0.80 |
| 30 | Basic | 0.90 | 0.79 | 0.89 | 0.78 |
| 30 | Detailed | 0.91 | 0.80 | 0.90 | 0.80 |
| Stratified sampling | | | | | |
| 3 | Basic | 0.92 | 0.84 | 0.83 | 0.72 |
| 3 | Detailed | 0.93 | 0.84 | 0.88 | 0.77 |
| 6 | Basic | 0.92 | 0.82 | 0.88 | 0.78 |
| 6 | Detailed | 0.92 | 0.81 | 0.89 | 0.78 |
| 9 | Basic | 0.90 | 0.78 | 0.88 | 0.78 |
| 9 | Detailed | 0.91 | 0.81 | 0.90 | 0.80 |
| 12 | Basic | 0.93 | 0.83 | 0.89 | 0.79 |
| 12 | Detailed | 0.92 | 0.82 | 0.89 | 0.78 |
| 15 | Basic | 0.89 | 0.79 | 0.89 | 0.79 |
| 15 | Detailed | 0.91 | 0.81 | 0.89 | 0.79 |
| 18 | Basic | 0.93 | 0.85 | 0.91 | 0.82 |
| 18 | Detailed | 0.93 | 0.85 | 0.91 | 0.80 |
| 21 | Basic | 0.92 | 0.83 | 0.89 | 0.79 |
| 21 | Detailed | 0.92 | 0.82 | 0.90 | 0.80 |
| 24 | Basic | 0.9 | 0.78 | 0.90 | 0.80 |
| 24 | Detailed | 0.92 | 0.81 | 0.90 | 0.81 |
| 27 | Basic | 0.92 | 0.83 | 0.89 | 0.79 |
| 27 | Detailed | 0.92 | 0.83 | 0.91 | 0.81 |
| 30 | Basic | 0.91 | 0.79 | 0.91 | 0.82 |
| 30 | Detailed | 0.91 | 0.79 | 0.91 | 0.81 |
| Zero-shot | | | | | |
| 0 | Basic | 0.80 | 0.69 | 0.81 | 0.70 |
| 0 | Detailed | 0.76 | 0.64 | 0.87 | 0.78 |

**Table C3.**
*F1 scores of Llama-3 models*

| Shot # | Prompt Template | Llama-3-70B-Instruct | | Llama3-8B-Instruct | |
|---|---|---|---|---|---|
| | | Weighted F1 Score | Macro F1 Score | Weighted F1 Score | Macro F1 Score |
| Random sampling | | | | | |
| 3 | Basic | 0.94 | 0.86 | 0.90 | 0.79 |
| 3 | Detailed | 0.95 | 0.88 | 0.93 | 0.83 |
| 6 | Basic | 0.94 | 0.85 | 0.91 | 0.80 |
| 6 | Detailed | 0.94 | 0.86 | 0.91 | 0.80 |
| 9 | Basic | 0.94 | 0.86 | 0.91 | 0.82 |
| 9 | Detailed | 0.94 | 0.86 | 0.92 | 0.83 |
| 12 | Basic | 0.94 | 0.86 | 0.92 | 0.82 |
| 12 | Detailed | 0.94 | 0.85 | 0.92 | 0.81 |
| 15 | Basic | 0.94 | 0.86 | 0.92 | 0.84 |
| 15 | Detailed | 0.94 | 0.86 | 0.92 | 0.83 |
| 18 | Basic | 0.94 | 0.85 | 0.9 | 0.79 |
| 18 | Detailed | 0.94 | 0.85 | 0.91 | 0.79 |
| 21 | Basic | 0.94 | 0.85 | 0.91 | 0.79 |
| 21 | Detailed | 0.94 | 0.85 | 0.90 | 0.78 |
| 24 | Basic | 0.94 | 0.86 | 0.91 | 0.81 |
| 24 | Detailed | 0.94 | 0.86 | 0.90 | 0.78 |
| 27 | Basic | 0.96 | 0.89 | 0.92 | 0.81 |
| 27 | Detailed | 0.96 | 0.89 | 0.92 | 0.82 |
| 30 | Basic | 0.95 | 0.86 | 0.91 | 0.80 |
| 30 | Detailed | 0.94 | 0.84 | 0.91 | 0.80 |
| Stratified sampling | | | | | |
| 3 | Basic | 0.95 | 0.87 | 0.91 | 0.81 |
| 3 | Detailed | 0.94 | 0.87 | 0.92 | 0.82 |
| 6 | Basic | 0.95 | 0.87 | 0.94 | 0.87 |
| 6 | Detailed | 0.95 | 0.87 | 0.92 | 0.83 |
| 9 | Basic | 0.94 | 0.86 | 0.93 | 0.86 |
| 9 | Detailed | 0.94 | 0.86 | 0.94 | 0.86 |
| 12 | Basic | 0.94 | 0.85 | 0.93 | 0.84 |
| 12 | Detailed | 0.94 | 0.85 | 0.92 | 0.82 |
| 15 | Basic | 0.95 | 0.87 | 0.94 | 0.87 |
| 15 | Detailed | 0.94 | 0.86 | 0.93 | 0.85 |
| 18 | Basic | 0.95 | 0.86 | 0.93 | 0.85 |
| 18 | Detailed | 0.94 | 0.85 | 0.93 | 0.84 |
| 21 | Basic | 0.95 | 0.88 | 0.93 | 0.85 |
| 21 | Detailed | 0.95 | 0.87 | 0.92 | 0.83 |
| 24 | Basic | 0.95 | 0.88 | 0.94 | 0.86 |
| 24 | Detailed | 0.95 | 0.87 | 0.93 | 0.85 |
| 27 | Basic | 0.95 | 0.88 | 0.93 | 0.86 |
| 27 | Detailed | 0.95 | 0.87 | 0.93 | 0.85 |
| 30 | Basic | 0.94 | 0.86 | 0.94 | 0.86 |
| 30 | Detailed | 0.95 | 0.87 | 0.93 | 0.83 |
| Zero-shot | | | | | |
| 0 | Basic | 0.93 | 0.83 | 0.90 | 0.80 |
| 0 | Detailed | 0.94 | 0.87 | 0.90 | 0.81 |

**Table C4.**
*F1 scores of Flan-UL2.*

| Shot # | Prompt Template | Weighted F1 Score | Macro F1 Score |
|---|---|---|---|
| Random sampling | | | |
| 3 | Basic | 0.90 | 0.79 |
| 3 | Detailed | 0.90 | 0.81 |
| 6 | Basic | 0.90 | 0.80 |
| 6 | Detailed | 0.90 | 0.79 |
| 9 | Basic | 0.91 | 0.81 |
| 9 | Detailed | 0.90 | 0.80 |
| 12 | Basic | 0.89 | 0.78 |
| 12 | Detailed | 0.89 | 0.78 |
| 15 | Basic | 0.91 | 0.79 |
| 15 | Detailed | 0.90 | 0.82 |
| Stratified sampling | | | |
| 3 | Basic | 0.90 | 0.80 |
| 3 | Detailed | 0.90 | 0.80 |
| 6 | Basic | 0.90 | 0.81 |
| 6 | Detailed | 0.90 | 0.81 |
| 9 | Basic | 0.90 | 0.80 |
| 9 | Detailed | 0.90 | 0.80 |
| 12 | Basic | 0.89 | 0.79 |
| 12 | Detailed | 0.89 | 0.79 |
| 15 | Basic | 0.90 | 0.81 |
| 15 | Detailed | 0.89 | 0.80 |
| Zero-shot | | | |
| 0 | Basic | 0.92 | 0.83 |
| 0 | Detailed | 0.92 | 0.82 |

***Appendix D.*** *Ill-formatted output categories in in-context learning results*

| | | |
|---|---|---|
| Missing initial labels | Irrelevant stances | Dual stances |
| Incorrect initial labels | Misindexing | Apologies or hallucinations |
| Empty responses | Creating new stance | Infinite repetitions |
| Task restatements | No label | |